\documentclass{article}

\usepackage[preprint,nonatbib]{neurips_2025}

\usepackage[utf8]{inputenc} 
\usepackage[T1]{fontenc}    
\usepackage{hyperref}       
\usepackage{url}            
\usepackage{booktabs}       
\usepackage{amsfonts}       
\usepackage{nicefrac}       
\usepackage{microtype}      
\usepackage{xcolor}         
\usepackage{subcaption}
\usepackage{booktabs} 
\usepackage{graphicx}
\usepackage{amsmath}
\usepackage{amssymb}
\usepackage{mathtools}
\usepackage{amsthm}
\usepackage{wrapfig}

\usepackage{booktabs}
\usepackage{multirow}
 \usepackage{colortbl} 
 \usepackage{ulem}

\newtheorem{theorem}{Theorem}[section]

\newtheorem{definition}{Definition}[section]

\newtheorem{observation}[theorem]{Observation}

\title{Walking the Tightrope: Disentangling Beneficial and Detrimental Drifts in Non-Stationary Custom-Tuning}

\author{Xiaoyu Yang, Jie Lu \thanks{Corresponding Author}, En Yu \\
Australian Artificial Intelligence Institute (AAII), \\ 
Faulty of Engineering and Information Technology,\\
University of Technology Sydney, Australia.\\
\texttt{xiaoyuyang386@gmail.com} 
}

\begin{document}

\maketitle

\begin{abstract}

This paper uncovers a critical yet overlooked phenomenon in multi-modal large language models (MLLMs): detrimental concept drift within chain-of-thought (CoT) reasoning during non-stationary reinforcement fine-tuning (RFT), where reasoning token distributions evolve unpredictably, thereby introducing significant biases in final predictions. 
To address this, we are pioneers in establishing the theoretical bridge between concept drift theory and RFT processes by formalizing CoT's autoregressive token streams as non-stationary distributions undergoing arbitrary temporal shifts. 
Leveraging this framework, we propose a novel counterfact-aware RFT that systematically decouples beneficial distribution adaptation from harmful concept drift through concept graph-empowered LLM experts generating counterfactual reasoning trajectories. 
Our solution, Counterfactual Preference Optimization (CPO), enables stable RFT in non-stationary environments, particularly within the medical domain, through custom-tuning of counterfactual-aware preference alignment. 
Extensive experiments demonstrate our superior performance of robustness, generalization and coordination within RFT. 
Besides, we also contributed a large-scale dataset CXR-CounterFact (CCF), comprising 320,416 meticulously curated counterfactual reasoning trajectories derived from MIMIC-CXR. Our code and data are public at: \url{https://anonymous.4open.science/r/CPO-FD61/}.
    
\end{abstract}

\section{Introduction}

Reinforcement Fine-Tuning (RFT) \cite{trungReFTReasoningReinforced2024,liuVisualRFTVisualReinforcement2025} has emerged as a promising paradigm for domain-specific customization of multi-modal large language models (MLLMs) \cite{alayracFlamingoVisualLanguage2022,baiQwenvlVersatileVisionlanguage2023,daiInstructBLIPGeneralpurposeVisionLanguage2023}, demonstrating remarkable capability in facilitating efficient domain shift with minimal data requirements, particularly for medical downstream tasks. However, the reinforcement-driven custom-tuning is fundamentally challenged by non-stationary environmental dynamics, especially for inherent domain-specific data characteristics such as long-tailed distributions in medical diagnosis, and systemic data imperfections including noise and sparsity. This complex synergy induces latent concept drift that progressively disaligns the model's representation space from domain reality, culminating in catastrophic error propagation that particularly jeopardizes the reliability of MLLMs in safety-critical applications like radiology report generation.


Concept drift theory \cite{luLearningConceptDrift2019,yang2025adapting} provides a new perspective for analyzing the domain shift of RFT in non-stationary custom-tuning, which focuses on the unpredictable distribution changes in data streams. We posit that the autoregressive decoding paradigm inherent to MLLMs can be characterized as a sequential token stream generation process. Within this framework, each token generation step propagates through the model's internal reasoning pathways, which remain opaque to external observation, while manifesting inherent stochasticity in the evolving token probability distributions across successive decoding iterations.

Within the concept drift framework, our analysis reveals critical limitations in reinforcement fine-tuning approaches that depend on verifiable rewards in chain-of-thought (CoT) \cite{liuVisualRFTVisualReinforcement2025}:
        
\begin{observation}
\label{ob.cd}
Specifically, while RFT operate through optimal reasoning pathway selection to maximize outcome certainty, we empirically observe that MLLM-generated CoT processes in specialized domains frequently demonstrate susceptibility to concept drift. This progressive deviation in intermediate reasoning ultimately induces substantial output divergence under non-stationary environments. 
\end{observation}

\begin{wrapfigure}{r}{0.4\textwidth}
        \includegraphics[width=0.4\textwidth]{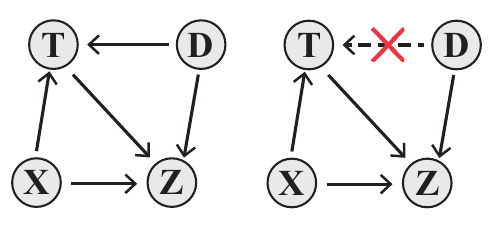}
        \caption{\textbf{Concept Drift in RFT's reasoning for chest diagnosis.} Despite analogous occurrence probabilities of "lung opacity" (in red) and "opacity" (in blue) tokens during the CoT, non-stationarity induces significant bad distributional drift in clinical conclusions, especially the opposite diagnosis of atelectasis, cardiomegaly and pneumonia.}
        \label{fig:toy-concept}
\end{wrapfigure}

Intuitively, we provide a representative case study that demonstrates this phenomenon in clinical reasoning contexts as presented in Fig.\ref{fig:toy-concept}. When diagnosing chest DR images, the model generates a reasoning trajectory containing the statement: "Asymmetric lung opacity in the right middle lobe is concerning for pneumonia." We found that in the token-level probability, "lung opacity" shows negligible differentiation from its ambiguous counterpart "opacity". Despite this minimal probabilistic disparity in the thinking process, our diagnostic outcome distribution analysis presents a radical divergence in final predictions as exhibited in Fig.\ref{fig:toy-concept}. In particular,  the diagnosis is completely opposite in terms of atelectasis, cardiomegaly and pneumonia. 

Therefore, summarizing the above challenges of reinforced fine-tuning in MLLMs, it raises the important question:

\textbf{\textit{How to adapt thinking process to concept drift under non-stationary reinforced custom-tuning?}}

Inspired by causal inference \cite{pearl1995causal,pearl2016causal,yangCausalInformedContrastiveLearning2025}, we develop Counterfactual Preference Optimization (CPO), a principled approach that systematically perturbs reasoning trajectories to discriminate between beneficial distribution adaptation and detrimental concept drift.

Firstly, we construct a hierarchical concept graph that codifies domain-specific knowledge structures through triadic relation embeddings, including positive correlation, irrelevance, and opposition. 
Subsequently, we structurally embed the hierarchically structured concept graph into the LLM's reasoning architecture as an expert-guided module, automatically generating semantically-constrained counterfactual inference paths. 
Consequently, during reinforced custom-tuning, we formulate a dual-preference optimization objective that jointly maximizes likelihood alignment with human preferences while minimizing similarity to adversarially generated counterfactual paths, thus achieving decoupling of beneficial domain adaptation from detrimental concept drift.
Finally, we contribute CXR-CounterFact (CCF), the chest diagnosis preference dataset comprising 320,416 fine-curated counterfactual reasoning trajectories derived from MIMIC-CXR \cite{johnson2019mimic} radiologic findings, aiming to validate our method and catalyze research advancements in counterfactual-aware reinforcement fine-tuning paradigms  

In summary, our paper mainly makes the following contributions:
\begin{enumerate} 
    \item First, we establish a novel theoretical framework that formalizes autoregressive token generation in MLLMs through the lens of concept drift theory, enabling systematic identification and causal analysis of detrimental reasoning divergence during non-stationary reinforced custom-tuning.
    \item Second, we propose Counterfactual Preference Optimization (CPO), which synergises structured domain-specific knowledge with systematic counterfactual intervention, driving the MLLMs with preference-aligned reinforcement learning. By embedding learnable concept graphs as the expert and generating adversarially-constrained reasoning trajectories, our approach achieves substantial decoupling between beneficial distribution adaptation and detrimental concept drift.
    \item Third, we conduct comprehensive empirical validation across various clinical benchmarks for chest radiograph, including disease classification, diagnostic report generation and zero-shot generalization. The superior results demonstrate statistically significant improvements in robustness, generalization, and accuracy of our method under non-stationary custom-tuning. Besides, we also provide ablation experiments to validate the effectiveness of various modules.
    \item As a pioneer contribution to the community, we introduce CXR-CounterFact (CCF), a large-scale dataset comprising 320,416 meticulously curated counterfactual reasoning trajectories derived from MIMIC-CXR. 
\end{enumerate}

\section{Methodology}

\begin{figure}[htbp]
    \centering
    \begin{subfigure}[t]{0.55\textwidth}
        \centering
        \includegraphics[height=0.15\textheight]{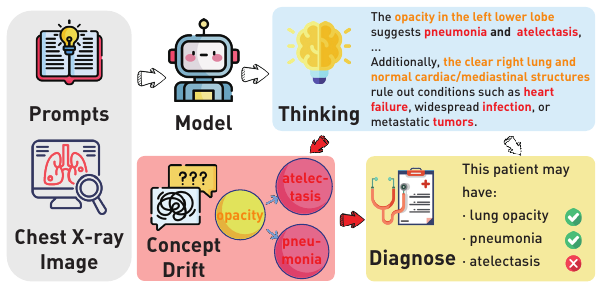}
        \caption{Concept Drift Behind the Thinking of MLLMs}
        \label{fig:CDinThk}
    \end{subfigure}
    \begin{subfigure}[t]{0.43\textwidth}
        \centering
        \includegraphics[height=0.15\textheight]{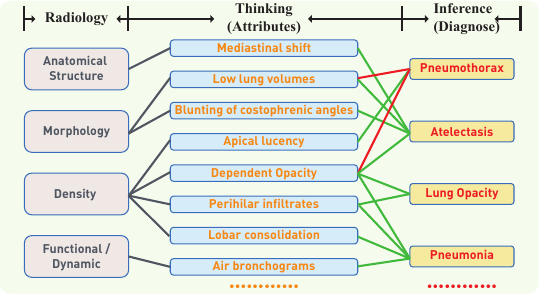}
        \caption{Concept Graph for Counterfactual Cause}
        \label{fig:KnowledgeTree}
    \end{subfigure}

    \begin{subfigure}[t]{0.99\textwidth}
        \centering
        \includegraphics[width=0.99\textwidth,height=0.16\textheight]{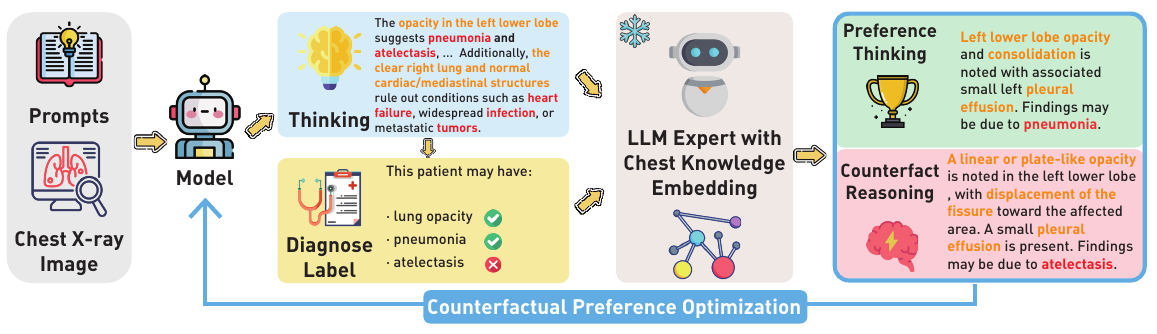}
        \caption{Counterfactual Preference Optimization}
        \label{fig:imbalanced}
    \end{subfigure}
    \caption{The main contributions of our methods. (a) 
    By formalizing autoregressive CoT generation as a stream of next-token prediction actions under the theoretical lens of concept drift, we reveal that even minor perturbations in reinforced fine-tuning can induce unpredictable distributional changes of final predicted results. (b) To disentangle detrimental drift, we introduce the concept graph that generates radiologically plausible counterfactual CoTs through controlled attribute perturbations.
    Green lines represent attributes that are positively correlated with the disease, while red denote they are exclusive. (c) We propose counterfactual preference optimization to drive the reinforced custom-tuning of MLLMs, enabling generalized CoT reasoning in non-stationary environments through disentanglement of beneficial domain adaptation from spurious concept drift, thereby achieving robust human-aligned decision-making via preference distillation.
    }
    \label{fig:analysis}
\end{figure}

\subsection{Underlying Concept Drift Behind Thinking}
\label{sec:2.1}

In this section, we first extend the concept drift theory to reinforced custom-tuning, highlighting the phenomenon wherein the distributional characteristics of targets undergo arbitrary changes over the course of the thinking process. 
Operating through recursive on-policy sampling, the MLLM $\pi$ autoregressively generates the token at position $j$ in the reasoning chain, conditioned on both the visual input image $v$, the textual prompt $l$, and the partial token sequence $t_{<j}$ of the CoT trajectory:
\begin{equation}
    t_{j} \sim \pi(\cdot | v, l, t_{<j})
\end{equation}

Therefore, we formally define concept drift behind the thinking as follows:
\begin{definition}
\label{defin}
The MLLM's autoregressive reasoning trajectory manifests as a thinking stream $S_{0,i} = \{s_{0},...,s_{i}\}$, where each cognitive state  $s_{j}=(t_{<j},z_{j})$  encapsulates all tokens generated so far $t_{<j}$ and its latent predicted distribution $z_{j}$ of the results by $t_{<j}$. Therefore, in position $i$, $S_{0,i}$ follows a certain distribution $F_{0,i}(x,z)$, thus the concept drift behind the thinking can be formalized as: 
\begin{equation}
    \label{eq:dis}
    \exists i: P_{i}(t,z) \neq P_{i+1}(t,z)
\end{equation}
where the joint probability $P_{i}(t,z)$ can be decomposed as $P_{i}(t,z) = P_{i}(t)\times P_{i}(z | t)$. 
\end{definition}
Consequently, this concept drift framework behind the thinking of MLLMs enables simultaneous characterization of temporal dynamics in Chain-of-Thought reasoning, formalized as the concept drift process $P_{i}(t)$, and its induced probabilistic divergence $P_{i}(z | t)$, capturing the evolving discrepancy between intended and actual outcome distributions throughout cognitive progression.

To adapt the reinforced custom-tuning to concept drift behind the CoT, it is essential to adapt the model to align with the evolving thinking distribution under non-stationary environment, which can be formally defined as:
\begin{equation}
\label{eq.goal}
    \min_{\pi^{(i)},\pi^{(i+1)},...,\pi^{(i+\tau)}} \sum_{i}^{i+\tau} \mathcal{L}( \prod_{j=1}^{L} \pi^{(i)}(t^{(i)}_{j} | v, l, t^{(i)}_{<j}) , y^{(i)}),
\end{equation}
where $\prod_{j=1}^{L} \pi^{(i)}(t^{(i)}_{j} | v, l, t^{(i)}_{<j})$ denotes the probability of CoT token sequence, $y^{(i)}$ represents the preferred CoT, $L$ symbolics the max length of tokens within the CoT, and $\pi^{(i)}$ signifies the MLLM at the cognitive status $i$. And the model is driven by the target metric $\mathcal{L}$ continuously to adapt the drift in a given time period $[i, i+\tau]$. Thus, we get the optimization object within the concept drift framework.



\subsection{Disentangling Concept Drift with Counterfactual Causes}
\label{sec.2.2}

\begin{wrapfigure}{r}{0.4\textwidth}
    \includegraphics[width=0.4\textwidth]{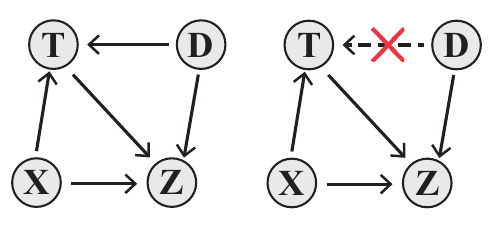}
    \caption{\textbf{Structural Causal Graph.}  \textbf{X}: Inputs, \textbf{Z}: Prediction Results, \textbf{T}: Chain-of-Thought, and \textbf{D}: Latent Concept Drift within CoT under Non-stationary Reinforced Custom-Tuning.}
    \label{fig:causal}
\end{wrapfigure}

The optimization objective formalized Eq.\ref{eq.goal} necessitates disentanglement of two competing goals: advantageous policy-induced domain adaptation and versus pathological concept drift arising from suboptimal policy execution, which are both sampled from policy $\pi$ within the time period $[i, i+\tau]$. However, it is challenging to determine the optimal preferred CoT and explicitly judge which strategies will cause unpredictable changes in tokens.

Fortunately, counterfactual causes provide an explicit manner to decouple these two competing goals. We construct a structural causal graph \cite{pearl1995causal,pearl2016causal} to formula the causal relationship among elements as discussed in Section \ref{sec:2.1}, including inputs ($X$) consisting of image $v$ and prompt $l$, prediction result ($Z$), Chain-of-Thought ($T$) and the concept drift in the reinforcement custom-tuning ($D$) as illustrated in Fig.\ref{fig:causal}, where $A\rightarrow B$ denotes that $A$ is the causer of $B$. The causal graph of $\{X,Z,T,D\}$ presents the following causal connections:

$(X,D)\rightarrow T$: This link denotes the chain-of-thought $T$ derived from the inputs $X$ through policy $\pi$ is under the impact of latent concept drift $D$.

$(X,T,D)\rightarrow Z$: This link presents that, apart from the regular reasoning pathway of$(X, T) \rightarrow Z$, the prediction is also impacted by the concept drift $D$ through the pathway of $D\rightarrow T\rightarrow Z$. 


In the constructed structural causal model,  nodes $D$ and $T$ are formally characterized as the confounder and mediator \cite{pearl2022direct}, respectively. The confounding variable  $D$
induces bias through the backdoor path  $X\leftarrow D\rightarrow Z$ \cite{pearl1995causal}, simultaneously influencing both the mediator $T$ and the outcome variable $Z$. This interference systematically distorts the estimation of CoT reasoning's causal effect on model predictions, particularly under non-stationary adaptation scenarios. The resulting spurious correlations manifest as concept drift artifacts that propagate through the mediation pathway  $X\rightarrow T \rightarrow Z$, ultimately compromising the stability of customized model tuning.

Building upon the above analysis grounded in cause, we formally decouple the concept drift dynamics in Chain-of-Thought reasoning of Eq.\ref{eq.goal} grounded in the cause. By constructing interventional distributions through $\textbf{do}$ operations, $P(Z| \textbf{do}(T=t), D=d)$, we quantify the latent causal effect:
\begin{equation}
\label{Eq.4}
    \psi = \mathbb{E}[Z_{T\leftarrow t, D\leftarrow d} - Z_{T\leftarrow t', D\leftarrow d}]
\end{equation}
where the potential outcome $T\leftarrow t$  represents the counterfactual scenario when forcibly maintaining the mediator $T$ at value $t$, while preserving the confounder state 
$d$. This formulation explicitly isolates the front-door effect $X\rightarrow T \rightarrow Z$ from backdoor concept drift propagation $X\leftarrow D \rightarrow Z$.

\subsection{Embedding Counterfactual Causes with LLM expert}
\label{sec:2.3}

Having operationalized concept drift decoupling through controlled counterfactual interventions in Section \ref{sec.2.2}, we identify the generation of autonomous counterfactual causes as the subsequent bottleneck that requires maintaining causal consistency within the chain-of-thought while avoiding semantic entanglement.

Accordingly, inspired by \cite{zhangKnowledgeenhancedVisuallanguagePretraining2023,zhouKnowledgeenhancedVisualLanguagePretraining2024}, we constructed a hierarchical concept graph for custom-tuning through autonomous knowledge extraction from chest radiograph reports. It systematically organizes medical concepts into four semantic dimensions: disease entities, radiographic features, clinical relationships, and taxonomies.
Specifically, the knowledge extraction pipeline leverages Med-PaLM \cite{singhalLargeLanguageModels2023a,singhalExpertlevelMedicalQuestion2025}, a medical domain-adapted large language model with radiological prior knowledge, to process 160,208 chest X-ray reports from the MIMIC-CXR dataset \cite{johnson2019mimic}. 
Through iterative semantic parsing, the model autonomously identifies 12 distinct pulmonary disease entities accompanied by 53 clinically relevant attributes. These attributes are meticulously annotated across four diagnostic categories, namely morphological alterations,  density anomalies, anatomical deviations, and functional/dynamic indicators.
To capture clinical interdependencies, we formalize three types of ontological relationships, including association, irrelevance, and exclusion. For instance, the framework automatically detects pathophysiological contradictions between emphysema-associated hyperinflation and atelectasis-related lung volume reduction, leading to the exclusion relationship between emphysema and atelectasis. As a toy example illustrated in Fig.\ref{fig:KnowledgeTree}, the resulting concept graph provides multi-relational representations where each disease entity is instantiated with its associated attributes and constraint relationships, enabling structured reasoning about pulmonary pathology.

Through systematic integration of the concept graph, Med-PaLM evolves as a customized LLM expert with causal prior knowledge of chest radiology, which effectively simulates radiologists' differential diagnosis protocols, and synthesises counterfactual diagnostic narratives through controlled feature perturbation while maintaining radiological plausibility, as exhibited below:

\begin{table}[htbp]
\centering
\begin{tabular}{
>{\columncolor[HTML]{EBEBEB}}p{13.5cm} }
\hline
\textbf{Ground Truth}                                                                                                                                                                                                                                                                                                                                                                                                                         \\
{\color[HTML]{0563C1} \textbf{Findings:}}                                                                                                                                                                                                                                                                                                                                                                                                                                 \\
Moderate cardiomegaly is increased. \uline{No focal consolidation or pneumothorax.}  There is a slight blunting of the costophrenic angles, which may indicate small pleural effusion or scarring. There is increased density at the perihilar regions which may indicate pulmonary vascular congestion.                                                                             \\
{\color[HTML]{0563C1} \textbf{Diagnosis:}}                                                                                                                                                                                                                                                                                                                                                                                                                                  \\
The disease of this patient is   Cardiomegaly.                                                                                                                                                                                                                                                                                                                                                                                          \\ \hline
\textbf{Generated Counterfactual Reasoning of Pneumonia.}                                                                                                                                                                                                                                                                                                                                                                                      \\ 
{\color[HTML]{0563C1} \textbf{Findings:}}                                                                                                                                                                                                                                   \\
PA and lateral views of the chest show moderate cardiomegaly. \uline{Focal consolidation is noted in the right lower lobe with accompanying bronchial airspace opacification.} No pneumothorax is observed. Slight blunting of the costophrenic angles suggests the presence of a small pleural effusion. \uline{Increased density at the perihilar regions indicates pulmonary vascular congestion, but also suggests possible pneumonia.} \\
{\color[HTML]{0563C1} \textbf{Diagnosis:}}                                                                                                                                                                                                                                                                                                                                                                                                                                \\
The disease of this patient is   Pneumonia.                                                                                                                                                                                                                                                                                                                                                                                             \\ \hline
\end{tabular}
\caption{Example to illustrate the generated counterfactual diagnosis. Ground Truth denotes the original diagnosis report from MIMIC-CXR, and the bottom is the counterfactual report designed for pneumonia. Underline indicates generated counterfactual diagnostic attributes through controlled perturbation while maintaining radiological plausibility, leading to the counterfactual diagnosis.}
\label{tab:my-table}
\end{table}

\subsection{Reinforced Custom-Tuning with Counterfactual Thinking}

Obtaining counterfactual diagnosis in Section.\ref{sec:2.3}, we propose Counterfactual Preference Optimzation (CPO) to drive the reinforced custom-tuning of the multi-modal large language models.

Formally, we have decomposed the CoT generation process into a stream of next token prediction actions $S_{0,i} = \{s_{0},...,s_{i}\}$, where each cognitive state $s_{j}=(t_{<j},z_{j})$  encapsulates all tokens generated so far $t_{<j}$ and its latent predicted distribution $z_{j}$ of the results by $t_{<j}$, as exhibited in Definition \ref{defin}. At timestep $j$, the action $t_{j}$ is sampled from the policy $\pi(\cdot | v, l, t_{<j})$ where $t_{j}$ can be any token in the vocabulary. After each action, the resulting state $s_{j+1}$ is the concatenation  of the current state $s_{j}$ and the action $t_{j}$ with its latent predicted results:
\begin{equation}
        s_{j+1}=(t_{<j} \circ t_{j}, P_{j}(z | t_{<j} \circ t_{j})), 0\leq j \leq L
\end{equation}
where $\circ$ denotes the concatenation between tokens stream $t_{<j}$ and action token $t_{j}$, $L$ represents the maximum length of CoT, and $P$ is the latent predicted distribution of results derived by $t_{<j}$ as presented in Eq.\ref{eq:dis}. As the start of the chain-of-thought, $a_{0}$ is usually the token <think>. While it produces the </think> token, the resulting state $s_{L+1}$ is the terminal state, thereby concluding one chain-of-thought generation process. Thereby, the chain-of-thought preferred by humans is considered to be $t^{+}$, namely the diagnosis report stemming from the radiologist, while the generated counterfactual CoT is represented by $t^{-}$.

Consequently, following the DPO \cite{rafailovDirectPreferenceOptimization2023}, we derive the optimal policy that maximizes the reward function through:
\begin{equation}
    \pi_{\theta}(t|v,l) \varpropto \pi_{\text{ref}}(t|v,l)\exp{(\frac{r(v,l,t)}{\beta})}
\end{equation}
where $\beta$ is a parameter controlling the deviation from the base reference policy $\pi_{\text{ref}}$, namely the initial supervised fine-tuned (SFT) model, and $\pi_{\theta}$ denotes the fine-tuning model. With counterfactual effect in Eq.\ref{Eq.4}, the reward difference between human-preferred positive samples and counterfactual samples can be defined as:
\begin{equation}
    r(v,l,t^{+}) - r(v,l,t^{-}) = \beta \left[ \log\frac{\pi_{\theta}(t^{+}|v,l)}{\pi_{\text{ref}}(t^{+}|v,l)} - \log\frac{\pi_{\theta}(t^{-}|v,l)}{\pi_{\text{ref}}(t^{-}|v,l)} \right]
\end{equation}
Thus, based on the Bradley-Terry model, the counterfactual preference optimization (CPO) is driving the reinforced custom-tuning of the MLLMs through the maximum likelihood objective:
\begin{equation}
    \mathcal{L}_{\text{CPO}} (\pi_{\theta};\pi_{\text{ref}}) = - \mathbb{E}_{(v,l,t^{+},t^{-})}\left[\log \sigma \left( \beta \log\frac{\pi_{\theta}(t^{+}|v,l)}{\pi_{\text{ref}}(t^{+}|v,l)} - \beta\log\frac{\pi_{\theta}(t^{-}|v,l)}{\pi_{\text{ref}}(t^{-}|v,l)} \right) \right]
\end{equation}

In this context, it culminates in counterfactual reinforced custom-tuning, an adaptive framework that effectively differentiates between advantageous domain adaptation and harmful concept drift in non-stationary environments, achieving equilibrium preservation through causal intervention and dynamic policy reinforcement, walking the tightrope.

\subsection{Building CXR-CounterFact Dataset for Clinical Reasoning Chains}

Since we are pioneers in introducing counterfactual cause into reinforced custom-tuning of MLLMs, we are deeply aware of the scarcity of counterfactual CoT in downstream tasks, especially in the highly professional medical field. Thus, our aspiration is for the model to adeptly acclimate to the concept drift by itself, acquiring abundant knowledge with more and more data, but not exhibiting bias.

In this context, a more realistic training dataset for multi-modal large language models is required to validate their potential to be trained under the non-stationary reinforced custom-tuning. Recognizing the demand for higher-quality multi-modal data with CoT, we develop a datasets called CXR-CounterFact Dataset (CCF), extending the MIMIC-CXR\cite{johnson2019mimic} with counterfactual chain-of-thought. This novel dataset introduces 320,416 meticulously curated counterfactual pairs spanning 14 thoracic pathologies, establishing a pioneering large-scale benchmark for causal interpretation in clinical chest X-ray analysis.
More details are given in Appendix \ref{appendxi:dataset}.

\section{Experiments}

In this section, we verify the robustness, generalization and coordination of our proposed counterfactual preference optimization in reinforced custom-tuning under non-stationary environments. 

\begin{wrapfigure}{r}{0.4\textwidth}
        \centering
        \includegraphics[width=0.4\textwidth]{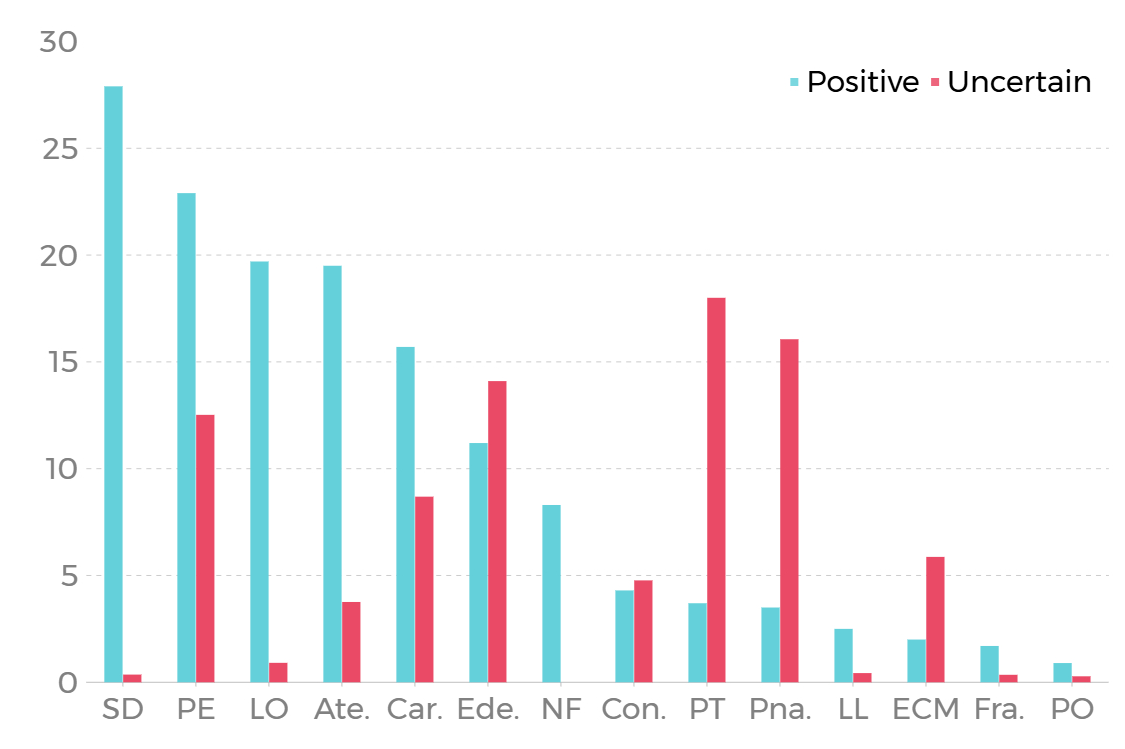}
        \caption{\textbf{Non-stationarity of MIMIC-CXR with its percentage of diseases.}    
        Blue signifies patients with clinically confirmed diagnoses showing the long-tailed characteristic, while red demarcates suspected cases emphasizing the inherent uncertainty within medicine.}
        \label{fig:mimic}
\end{wrapfigure}

MIMIC-CXR\cite{johnson2019mimic}  is utilized to train the MLLMs via reinforced custom-tuning for domain adaptation, which presents 371,920 chest X-rays associated with 227,943 imaging studies from 65,079 patients. And images are provided with 14 labels with corresponding free-text radiology reports, namely Atelectasis (Ate.), Cardiomegaly (Car.), Consolidation (Con.), Edema (Ede.), Enlarged Cardiomediastinum (ECM), Fracture (Fra.), Lung Lesion (LL), Lung Opacity (LO), Pleural Effusion (PE), Pneumonia (Pna.), Pneumothorax (Pnx.), Pleural Other (PO), Support Devices (SD) and No Finding (NF).

We selected the MIMIC-CXR \cite{johnson2019mimic} dataset not only for its well-established benchmark enabling rigorous performance evaluation on real-world downstream medical tasks, but also due to its authentic clinical representation that exhibits inherent non-stationarity, particularly long-tail and diagnostic ambiguity. As illustrated in Fig.\ref{fig:mimic}, the statistical profiling of 14 thoracic pathologies in MIMIC-CXR reveals dual clinical characteristics: the number of confirmed diseases showed a clear long-tail distribution, and each disease had a large number of uncertain patients. Beyond that, we found that 40.87\% of the patients suffered from two or more diseases, with nearly 19.97\% experiencing three or more. They all reflect the non-stationary environments of our experimental setup within MIMIC-CXR as the training dataset for reinforced custom-tuning. 

In terms of the model, we employ Qwen2.5-VL (7B) \cite{qwen2.5-VL} to perform supervised fine-tuning (SFT) and reinforced fine-tuning (RFT), cascadedly. And they only train one epoch with a batch size of 2.

More detailed experimental implementations are given in Appendix \ref{appendxi:exp}.

\subsection{Taming the Non-stationary Custom-Tuning}

\begin{table}[htbp]
\centering
\setlength{\tabcolsep}{1.5mm}{
\begin{tabular}{@{}llcccccc@{}}
\toprule
                                                            & Venue                        & Con.                                 & PE                                   & Pna.                                 & Pnx.                                 & Ede.                        & Avg.                                 \\ \midrule
CTrans \cite{Bannur_2023_CVPR}                   & CVPR'23                      & 44.0                                 & 61.3                                 & 45.1                                 & 31.5                                 & 65.5                        & 49.5                                 \\
CheXRelNet  \cite{karwande2022chexrelnet}                   & MICCAI'22                    & 47.0                                 & 47.0                                 & 47.0                                 & 36.0                                 & 49.0                        & 45.2                                 \\
BioViL \cite{boecking2022making}                            & ECCV'22                      & 56.0                                 & 63.0                                 & 60.2                                 & 42.5                                 & 67.5                        & 57.8                                 \\
BioViL-T    \cite{Bannur_2023_CVPR}                         & CVPR'23                      & 61.1                                 & 67.0                                 & 61.9                                 & 42.6                                 & 68.5                        & 60.2                                 \\
Med-ST \cite{yang2024unlocking}                             & ICML'24                      & 60.6                                 & 67.4                                 & 58.5                                 & 65.0                                 & 54.2                        & 61.1                                 \\
TempA-VLP \cite{10943948}                                   & WACV'25                      & 65.2                                 & 59.4                                 & {\color[HTML]{0000FF} 73.4}          & 43.1                                 & {\color[HTML]{FF0000} 77.1} & 63.6                                 \\
CoCa-CXR   \cite{chen2025cocacxrcontrastivecaptionerslearn} & Arxiv'25                     & {\color[HTML]{0000FF} 70.4}          & 69.6                                 & 61.4                                 & 72.8                                 & 71.8                        & 69.2                                 \\ \midrule
SFT                                                         &                              & 54.9                                 & {\color[HTML]{0000FF} 71.7}          & 70.0                                 & {\color[HTML]{FF0000} 95.9}          & {\color[HTML]{0000FF} 76.5} & {\color[HTML]{0000FF} 73.8}          \\
CPO                                                         & \multirow{-2}{*}{This paper} & {\color[HTML]{FF0000} \textbf{77.7}} & {\color[HTML]{FF0000} \textbf{72.7}} & {\color[HTML]{FF0000} \textbf{87.4}} & {\color[HTML]{0000FF} \textbf{95.8}} & \textbf{75.3}               & {\color[HTML]{FF0000} \textbf{81.8}} \\ \bottomrule
\end{tabular}
}
\caption{\textbf{Evaluation results of multi-label chest diseases classification on MS-CXR-T.} Top-1 accuracy is applied to evaluate the performance of different methods. The best-performing models are highlighted in red, with the second-best in blue. SFT denotes the results of supervised fine-tuning, while the CPO represents our counterfactual preference optimization method. }
\label{tab:MIMIC-CXR-T}
\end{table}

First, to explicitly demonstrate the superior performance of our proposed method under non-stationary environments, especially in robustness, we compare it with other models on MS-CXR-T \cite{bannurLearningExploitTemporal2023}, where instances are chosen from the public MIMIC-CXR. As exhibited in Table \ref{tab:MIMIC-CXR-T}, our counterfactual preference optimization approach achieve the superior overall performance of 81.8\%, surpassing the second CoCa-CXR \cite{chen2025cocacxrcontrastivecaptionerslearn} nearly 12.6\%. It demonstrates the robustness of our approach to reinforced fine-tuning under non-stationary environments. 
While our approach trails TempA-VLP \cite{10943948} by 1.8\% on edema (Ede.) detection, we argue that this performance gap emerges from the utilization of additional annotations from Chest Im-aGenome \cite{wu2021chest} in addition to standard MIMIC-CXR. In terms of the pneumothorax (Pnx.), SFT has achieved a high result of 95.9\%, so the slight decrease in CPO does not affect the overall performance of the model.

Notably, supervised fine-tuning  (SFT) exhibits suboptimal performance on the clinically correlated diseases, namely consolidation (Con.) and pneumonia (Pna.), as presented in Table \ref{tab:my-table} of Section.\ref{sec:2.3}. It empirically validates our Observation \ref{ob.cd} that the inherent concept drift in disease attribute representation within chain-of-thought reasoning introduces systematic prediction biases.

Beyond that, the substantial performance gains of 22.8\% and 17.4\% in CPO for  consolidation (Con.) and pneumonia (Pna.), respectively, underscore our core contribution, which disentangles the concept drift of CoT in reinforcement learning under non-stationary custom-tuning, achieving robust reasoning.

\subsection{Concept Drift-Aware CoT for Accurate Reasoning}

\begin{table}[htbp]
\centering
\setlength{\tabcolsep}{0.6mm}{
\begin{tabular}{@{}llccccccc@{}}
\toprule
                                                              & Venue               & BLEU-1                                & BLEU-2                                & BLEU-3                                & BLEU-4                                & ROUGE-L                               & METEOR                                & CIDEr                                 \\ \midrule
R2Gen \cite{chen2020generating}                               & EMNLP'20            & 0.353                                 & 0.218                                 & 0.145                                 & 0.103                                 & 0.277                                 & 0.142                                 & -                                     \\
PPKED \cite{liu2021exploring}                                 & CVPR'21             & 0.36                                  & 0.224                                 & 0.149                                 & 0.106                                 & 0.284                                 & 0.149                                 & 0.237                                 \\
AlignTrans \cite{you2021aligntransformer}                     & MICCAI'2            & 0.378                                 & 0.235                                 & 0.156                                 & 0.112                                 & 0.283                                 & 0.158                                 & -                                     \\
CMCL \cite{liu2021competence}                                 & ACL'21              & 0.344                                 & 0.217                                 & 0.14                                  & 0.097                                 & 0.281                                 & 0.133                                 & -                                     \\
Clinical-BERT \cite{yan2022clinical}                          & AAAI'22             & 0.383                                 & 0.230                                 & 0.151                                 & 0.106                                 & 0.275                                 & 0.144                                 & 0.151                                 \\
METransformer   \cite{wang2023metransformer}                  & CVPR'23             & 0.386                                 & 0.250                                 & 0.169                                 & 0.124                                 & {\color[HTML]{0000FF} 0.291}          & 0.152                                 & {\color[HTML]{0000FF} 0.362}          \\
DCL \cite{li2023dynamic}                                      & CVPR'23             & -                                     & -                                     & -                                     & 0.109                                 & 0.284                                 & 0.150                                  & 0.281                                 \\
R2GenGPT \cite{wang2023r2gengpt}                              & MetaRad'23    & 0.408                                 & 0.256                                 & 0.174                                 & 0.125                                 & 0.285                                 & 0.167                                 & 0.244                                 \\
PromptMRG \cite{jin2024promptmrg}                             & AAAI'24             & 0.398                                 & -                                     & -                                     & 0.112                                 & 0.268                                 & 0.157                                 & -                                     \\
BtspLLM \cite{liu2024bootstrapping}                           & AAAI'24             & 0.402                                 & 0.262                                 & 0.18                                  & 0.128                                 & {\color[HTML]{0000FF} 0.291}          & {\color[HTML]{0000FF} 0.175}          & -                                     \\
MambaXray   \cite{wangCXPMRGBenchPretrainingBenchmarking2024} & Arxiv'24            & {\color[HTML]{0000FF} 0.422}          & {\color[HTML]{0000FF} 0.268}          & {\color[HTML]{0000FF} 0.184}          & {\color[HTML]{0000FF} 0.133}          & 0.289                                 & 0.167                                 & 0.241                                 \\ \midrule
CPO                                                 & This paper & {\color[HTML]{FF0000} \textbf{0.426}} & {\color[HTML]{FF0000} \textbf{0.288}} & {\color[HTML]{FF0000} \textbf{0.186}} & {\color[HTML]{FF0000} \textbf{0.155}} & {\color[HTML]{FF0000} \textbf{0.321}} & {\color[HTML]{FF0000} \textbf{0.236}} & {\color[HTML]{FF0000} \textbf{0.375}} \\ \bottomrule
\end{tabular}
}
\caption{Evaluation results of diagnostic report generation on MIMIC-CXR with various metrics including BLEU-1/-2/-3/-4, ROUGE-L, METEOR and CIDEr. The best-performing models are highlighted in red, with the second-best in blue.}
\label{tab:CoT}
\end{table}

Beyond the classification, we verify our main contribution of accurate reasoning, preserving the beneficial CoT within domain adaptation, while eliminating harmful concept drift. As exhibited in the Table \ref{tab:CoT}, the experiments of diagnostic report generation on MIMIC-CXR are conducted to assess the performance of the thinking in our proposed model with chain-of-thought. Our evaluation combines multiple metrics: BLEU evaluates terminology accuracy with higher-order scores indicating logical coherence in clinical reasoning, ROUGE-L assesses completeness through narrative alignment, METEOR enables synonym-aware lexical matching, and CIDEr prioritizes clinical details via corpus-informed weighting.

The experimental findings demonstrate that our reasoning framework achieves prominent performance across all evaluation metrics, with particularly notable improvements in BLEU-4 (16.5\% improvement), ROUGE-L (10.3\% increase) and METEOR (34.8\% enhancement) scores, indicating the coherence, completeness, and professionalism of our model's thinking. We attribute it to the enhanced fidelity of our reasoning chains in combating concept drift during non-stationary reinforcement learning processes. 
These enhancements reveal that our method's superior accuracy stems from its capacity to maintain coherent reasoning pathways even when faced with dynamically shifting environmental parameters of complex RL scenarios.

\subsection{Generalized Reinforced Custom-tuning}

\begin{table}[htbp]
\centering
\begin{tabular}{@{}lcccccc@{}}
\toprule
Method                                & Open-I                               & PadChest                             & PadChest20                           & ChestXray14                          & ChestXpert                           & ChestXDet10                          \\ \midrule
MedCLIP \cite{wang2022medclip}        & 55.1                                 & 50.8                                 & 50.1                                 & 56.4                                 & 74.4                                 & 57.1                                 \\
BiomedCLIP \cite{zhang2023biomedclip} & 57.7                                 & 51.3                                 & 51.0                                 & 63.9                                 & 67.7                                 & 63.0                                 \\
GLoRIA \cite{huang2021gloria}         & 58.9                                 & 56.5                                 & 55.8                                 & 61.0                                 & 75.0                                 & 64.5                                 \\
BioViL \cite{Bannur_2023_CVPR}        & 70.2                                 & 65.5                                 & 60.8                                 & 72.9                                 & 78.9                                 & 70.8                                 \\
CheXzero \cite{tiu2022expert}         & 75.9                                 & 62.9                                 & 68.8                                 & 72.6                                 & 87.9                                 & 71.3                                 \\
MedKLIP \cite{wu2023medklip}          & 75.9                                 & 62.9                                 & 68.8                                 & 72.6                                 & 87.9                                 & 71.3                                 \\
KAD \cite{zhang2023knowledge}         & 80.7                                 & 75.0                                 & 73.5                                 & 78.9                                 & 90.5                                 & 73.5                                 \\
CARZero \cite{lai2024carzero}         & {\color[HTML]{0000FF} 83.8}          & {\color[HTML]{0000FF} 81.0}          & {\color[HTML]{0000FF} 83.7}          & {\color[HTML]{0000FF} 81.1}          & {\color[HTML]{0000FF} 92.3}          & {\color[HTML]{0000FF} 79.6}          \\ \midrule
CPO                                   & {\color[HTML]{FF0000} \textbf{84.4}} & {\color[HTML]{FF0000} \textbf{82.0}} & {\color[HTML]{FF0000} \textbf{85.1}} & {\color[HTML]{FF0000} \textbf{81.7}} & {\color[HTML]{FF0000} \textbf{92.5}} & {\color[HTML]{FF0000} \textbf{80.1}} \\ \bottomrule
\end{tabular}
\caption{\textbf{Evaluation results of zero-shot diseases classification on Open-I\cite{demner2012design}, PadChest\cite{bustos2020padchest}, PadChest20 \cite{bustos2020padchest}, ChestXray14 \cite{Wang_2017_CVPR}, ChestXpert \cite{irvin2019chexpert} and ChestXDet10 \cite{liu2020chestxdet10}.} AUC is applied to evaluate the performance of different methods. The best-performing models are highlighted in red, with the second-best in blue.}
\label{tab:Gen}
\end{table}

Furthermore, we validated the generalization of our model on downstream tasks with zero-shot multi-label classification across six different benchmarks, as presented in Table \ref{tab:Gen}. 
Experimental results demonstrate that our CPO-driven MLLMs achieve zero-shot superiority over the second-best baseline CARZero  \cite{lai2024carzero} across all benchmark datasets, underscoring our remarkable robustness and generalization capabilities even when trained under non-stationary environmental regimes.

\subsection{Ablation Study on Inherent Compatibility of CPO and CoT: Two Peas in a Pod}

\begin{wraptable}{r}{0.6\textwidth}
\centering
\setlength{\tabcolsep}{1.5mm}{
\begin{tabular}{@{}ccccccccc@{}}
\toprule
\multirow{2}{*}{SFT} & \multicolumn{2}{c}{RFT} & \multirow{2}{*}{Con.} & \multirow{2}{*}{PE} & \multirow{2}{*}{Pna.} & \multirow{2}{*}{Pnx.} & \multirow{2}{*}{Ede.} & \multirow{2}{*}{Avg.} \\
                     & CoT                    & CPO                   &                       &                     &                       &                       &                       &                       \\ \midrule
\checkmark           & -                      & -                     & 54.9                  & 71.7                & 70.0                  & 95.9                  & 76.5                  & 73.8                  \\
\checkmark           & \checkmark             & -                     & 58.4                  & 71.2                & 75.0                  & 94.4                  & 75.5                  & 74.9                  \\
\checkmark           & -                      & \checkmark            & 70.5                  & 72.7                & 77.3                  & 95.2                  & 75.8                  & 78.3                  \\
\checkmark           & \checkmark             & \checkmark            & 77.7                  & 72.7                & 87.4                  & 95.8                  & 75.3                  & 81.8                  \\ \bottomrule
\end{tabular}
}
\caption{\textbf{Ablation evaluation results on chain-of-thought (CoT) and counterfactual preference (CPO) within reinforced fine-tuning (RFT) on MIMIC-CXR}, where all RFT stages follow the supervised fine-tuning (SFT).
The \checkmark denotes that the results are trained with the corresponding module. The results are based on the test split of the MS-CXR-T, with Top-1 accuracy (Acc) as the metric.}
\label{tab:abl}
\end{wraptable}

Moreover, we conduct ablation experiments on MIMIC-CXR to validate the feasibility and coordination of the chain-of-thought (CPO) and counterfactual preference optimization (CPO) within reinforced fine-tuning (RFT) under non-stationary environments, as presented in Table \ref{tab:abl}. Among them, only CoT without CPO in RFT represents the utilization of direct preference optimization (DPO) \cite{rafailovDirectPreferenceOptimization2023} to drive MLLMs for reinforcement learning. While, the only CPO in RFT denotes the reinforced fine-tuning only applies the diagnosis results without the thinking process during the training.

The experimental analysis reveals CPO's superior performance gain in reinforcement learning (4.5\% vs. CoT's 1.1\%). We argue that it is mainly attributable to its mechanism of introducing causally attributed negative samples that enable decision boundary refinement in feature space, whereas CoT primarily operates through stepwise cognitive scaffolding via enhanced positive samples.Therefore, the inherent synergistic compatibility between CoT and CPO emerges through their complementary roles in reinforcement learning frameworks, with CoT generating reinforcement-aligned positive exemplars and CPO providing causality-attuned negative specimens, jointly orchestrating MLLM training optimization as empirically validated through comprehensive benchmarking results achieving state-of-the-art performance.

\section{Conclusion and Outlooks}

In this paper, we present counterfactual preference optimization (CPO), a novel, robust and generalized reinforced custom-tuning paradigm tailored for non-stationary environments. We employ concept drift theory to methodically formalize the bias within the autoregressive token generation of MLLMs and put forward a causal counterfactual thinking to mitigate these detrimental drifts and keep good domain adaptation. By virtue of this framework, CPO is devised to counteract the unpredictable distribution changes occurring within non-stationary environments.

We hope that our work will inspire future advancements in counterfactual cause of reinforced learning paradigm, specifically addressing biases originating from real-world data challenges. In future research, we will focus on the efficiency of counterfactual causes in reinforced fine-tuning.

\bibliographystyle{nips}
\bibliography{example_paper}

\medskip

\clearpage
\newpage
\appendix

\section*{Appendix}

\section{Related Works}
\label{appendix:relatedwork}

\subsection{Concept Drift}

In their survey spanning multiple studies, Lu et al. \cite{luLearningConceptDrift2019, lu2020data} establish a comprehensive taxonomy of concept drift mitigation strategies, categorizing prevailing approaches into three principal paradigms: error rate-driven adaptations \cite{wangSelfadaptiveEnsembleUser2024, jiaoDynamicEnsembleSelection2024}, data distribution-aware methodologies \cite{yang2025adapting,cerqueiraSTUDDStudentTeacher2023,yangTdistributedSphericalFeature2023}, and multi-hypothesis frameworks \cite{yu2024online, yuLearnadaptConceptDrift2022}. 
Our work aligns with the distribution-centric paradigm, which distinguishes itself through its dual capacity for both precise drift detection via rigorous statistical analysis and holistic drift characterization across temporal, spatial, and quantitative dimensions. Specifically, these distribution-driven techniques enable not merely the identification of concept drift occurrence but also facilitate granular diagnostics through temporal localization of drift emergence, feature subspace attribution, and severity quantification - capabilities that render them particularly advantageous for developing interpretable adaptive systems requiring both drift awareness and targeted model recalibration.

Recent advances in concept drift adaptation have yielded sophisticated methodologies across diverse learning scenarios. The Online Boosting Adaptive Learning (OBAL) framework \cite{yu2024online} has emerged as a dual-phase solution for multistream classification challenges, initially employing Adaptive Covariate Shift Adaptation (AdaCOSA) to model dynamic inter-stream correlations before transitioning to Gaussian Mixture Model-based weighting for asynchronous drift mitigation. Complementing this, CDMLLM \cite{yang2025adapting} reveals critical vulnerabilities in vision-language models through systematic analysis of concept drift-induced biases across pre-training and fine-tuning stages, proposing a unified framework that synergizes T-distribution adaptation for long-tailed calibration with explicit out-of-distribution detection to enhance multimodal alignment robustness. Expanding the scope beyond individual data streams, GDDM \cite{yuDetectingGroupConcept2023} introduces a distribution-free statistical framework for detecting subtle group-level concept drifts in multi-stream environments through adaptive hypothesis testing mechanisms. In parallel, DDG-DA \cite{liDDGDataDistributionGeneration2022} pioneers anticipatory concept drift adaptation by modeling environmental evolution through predictive factor analysis and synthetic data generation, effectively bridging current observations with projected distribution shifts. Advancing unsupervised detection paradigms, STUDD \cite{cerqueiraSTUDDStudentTeacher2023} establishes a teacher-student discrepancy framework that leverages predictive consistency analysis to enable label-agnostic drift identification while maintaining detection sensitivity, thereby addressing practical deployment constraints in evolving data environments.

\subsection{Causal Inference}

Recently, increasing researchers have incorporated causal inference into deep-learning models, especially in large models. Deconfounded Image Captioning (DIC) \cite{yangDeconfoundedImageCaptioning2023} is proposed to address dataset bias in vision-language models through a causal lens, that integrates backdoor and front-door adjustments for systematic bias mitigation. The framework provides principled causal analysis of spurious correlations in multimodal alignment, offering theoretical grounding for decomposing bias sources through structured interventions. Likewise, aiming for spurious correlations induced by visual and linguistic biases during training, CIIC \cite{liuShowDeconfoundTell2022} is proposed as a causal intervention framework combining an Interventional Object Detector (IOD) and Interventional Transformer Decoder (ITD) guided by structural causal models. By applying backdoor adjustment through IOD's feature disentanglement and ITD's dual de-confounding mechanism, their approach systematically mitigates confounding effects across encoding and decoding stages, demonstrating enhanced generalization through causal correlation modeling. Similarly, targeting multi-hop fact verification bias in the large language model, Causal Walk \cite{zhangCausalWalkDebiasing2024} is proposed, a front-door adjustment framework that disentangles complex spurious correlations in evidence chains. 
The method models reasoning paths as mediators in structural causal models, decomposing causal effects via random walk-based treatment-mediator estimation and geometric mean-based mediator-outcome approximation. By integrating adversarial and symmetric datasets synthesized with large language models, the approach demonstrates superior debiasing performance.

Recent advances in causal representation learning have produced innovative methodologies to address confounding biases in large models. The C2L framework \cite{choiC2LCausallyContrastive2022} tackles model fragility through contrastive counterfactual synthesis, introducing a collective decision mechanism that aggregates predictions across probabilistically generated counterfactual sets while enforcing causal invariance via distributional consensus supervision, thereby overcoming dataset-inherent bias limitations of conventional augmentation approaches. Building on causal interpretability, ABCD \cite{rohekarCausalInterpretationSelfAttention2023} establishes formal theoretical grounding for Transformer architectures by reinterpreting self-attention mechanisms as structural equation estimators that capture conditional independence relations through partial correlation analysis in deep attention layers, enabling zero-shot causal discovery over input sequences while accounting for latent confounders through repurposed pre-trained models. Expanding the causal intervention paradigm, Causal Attention (CATT) \cite{yangCausalAttentionVisionLanguage2021} implements front-door adjustment via dual-path processing of In-Sample and Cross-Sample Attention, strategically integrating external contextual information through CS-ATT while preserving standard attention mechanisms to dynamically mitigate spurious correlations without explicit confounder specification, thereby achieving bias-resistant vision-language alignment through implicit causal disentanglement. Moreover, ResilientCL \cite{yangCausalInformedContrastiveLearning2025,yangMaskedImageContrastive2024} proposes the causal interventional contrastive objective to mitigate the concept drift within the momentum network of contrastive pre-training paradigm.

\subsection{Reinforced Fine-tuning}

The integration of reinforcement learning (RL) into post-training alignment of large language models (LLMs) has undergone remarkable evolution since OpenAI's seminal work on Reinforcement Learning from Human Feedback (RLHF) \cite{christiano2017deep}, which established a foundational paradigm for aligning model outputs with human values \cite{ouyang2022training}. While early implementations like OpenAI-o1 \cite{jaech2024openai} demonstrated the efficacy of human preference modeling, the prohibitive costs of manual annotation have catalyzed a paradigm shift toward automated reward generation through pre-trained systems. This transition has yielded innovative methodologies ranging from Bai et al.'s \cite{bai2022constitutional} constitutional approach utilizing sparse natural language feedback as proxy signals, to DeepSeek's progressive framework that first established baseline performance through pure RL (R0) before introducing their R1 variant \cite{guo2025deepseek}. The latter achieved enhanced generalization through cyclic alternation between supervised fine-tuning and their novel GRPO optimization protocol \cite{shao2024deepseekmath}, exemplifying the field's progression toward self-contained alignment systems.

Besides, the landscape of alignment methodologies continues to diversify through innovative paradigms: ReST \cite{gulcehre2023reinforcedselftrainingrestlanguage} employs iterative self-generation of policy-derived samples to refine LLMs via offline reinforcement learning, while DPO \cite{rafailovDirectPreferenceOptimization2023} fundamentally reformulates alignment as direct preference optimization through implicit reward modeling. Concurrent developments span Rejection Sampling Fine-Tuning's \cite{yuan2024scaling} curation of validated reasoning trajectories for supervised augmentation, and ReFT's \cite{trungReFTReasoningReinforced2024} phased optimization combining SFT initialization with PPO-driven exploration of automated reasoning path generation. Building upon these foundations, Visual-RFT \cite{liuVisualRFTVisualReinforcement2025} extends GRPO-based strategies to multimodal contexts, enhancing visual-language alignment under data scarcity, whereas B-STaR \cite{zeng2025bstar} introduces dynamic configuration adaptation for self-teaching systems through principled exploration-exploitation balancing. Pushing the boundaries of evaluation rigor, Qwen-Math-PRM \cite{zhang2025lessons} synergizes Monte Carlo estimation with LLM-as-judge consensus filtering while pioneering a hierarchical assessment framework integrating stepwise and holistic performance metrics.
Moreover, ViLaM \cite{yang2024enhancing} performs visual grounding unsupervised via reinforced learning under the open-world environment.

\section{CXR-CounterFact (CCF) Dataset}
\label{appendxi:dataset}

\begin{figure}[htbp]
    \centering
    \includegraphics[width=0.9\textwidth]{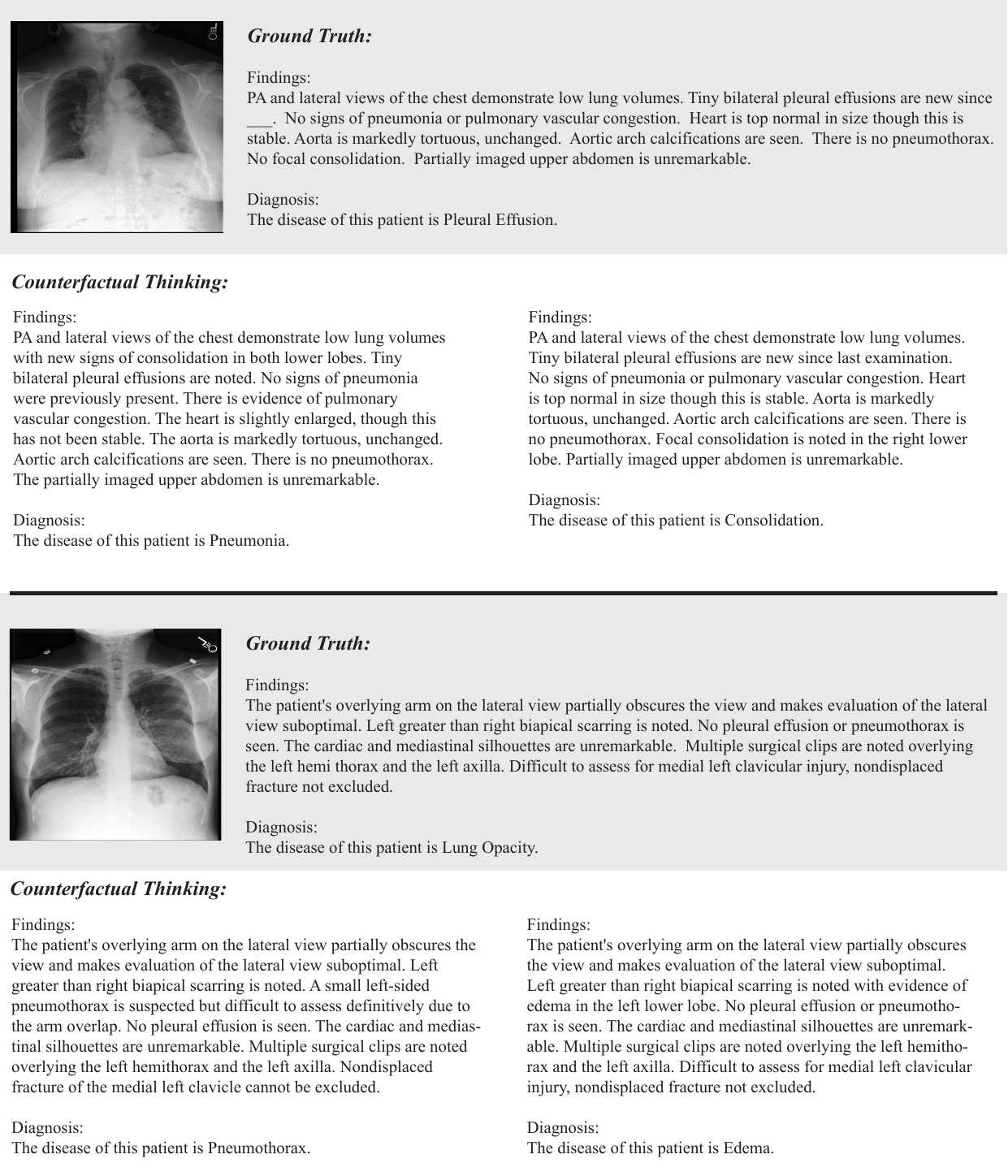}
    \caption{Samples of CXR-CounterFact (CCF) Dataset.}
    \label{fig:ccf}
\end{figure}

Figure \ref{fig:ccf} showcases the samples utilized for training and validation in our study. 
We use the  Med-PaLM \cite{singhalLargeLanguageModels2023a,singhalExpertlevelMedicalQuestion2025} to generate the related caption of the image, with the prompt of: 

\textit{"This is a radiology chest DR examination report of a patient: <Report>.}    

\textit{This is a diagram of the relationship between lung diseases and their radiographic manifestations:
<Concept Graph>}

\textit{Please generate a counterfactual radiology text showing <disease> based on the relationship and above context, with the same formatting."
}.

As depicted in Figure \ref{fig:ccf} , comprehensive descriptions of the image are provided through long-form text, encompassing details such as size, position, relationships, and other relevant information about the disease present in the image. This ensures a detailed and information-rich depiction of the visual content.
We have publicly released the datasets used for training and validation.

\section{Implementation Details}
\label{appendxi:exp}

In this section, implementation details are provided.

In terms of the supervised fine-tuning progress, the hyperparameters are presented in Table \ref{table:param}. Qwen2.5-VL (7B) \cite{qwen2.5-VL} is applied as our pre-trained model. During the SFT, we utilize the AdamW optimizer, which is configured with a cosine annealing schedule as the learning policy. The initial learning rate is set to $1\times10^{-4}$, and the AdamW optimizer is employed with hyperparameters $\beta= (0.9, 0.98)$. Additionally, we set the weight decay to 0.05 and the dropout rate to 0.1. During the first 20 warm-up steps, the learning rate increases to $1\times10^{-4}$, and subsequently decays to $10^{-7}$. Unless otherwise specified, the supervised fine-tuning of our multi-modal large language model consists of 660 steps, executed on $2\times 2$ NVIDIA A100 GPUs.

\begin{table}[htbp]
    \caption{The training hyperparameters of our MLLM.}
    \label{table:param}
    \centering
    \begin{minipage}[c]{0.45\textwidth}
        \setlength{\tabcolsep}{5mm}{
            \begin{tabular}{@{}lc@{}}
            \toprule
            \multicolumn{2}{l}{\textbf{Supervised Fine-tuning}} \\ \midrule
            Training Steps       & 660          \\
            Warmup Steps         & 20            \\
            Warmup Ratio         & 0.05         \\
            Optimizer            & AdamW        \\
            Learning Rate        & 1e-4         \\
            Learning Rate Decay  & Cosine       \\
            Adam $\beta$         & (0.9, 0.98)  \\
            Weight Decay         & 0.05         \\
            Batch Size           & 15           \\ \bottomrule
            \end{tabular}
            }        
    \end{minipage}
    \begin{minipage}[c]{0.45\textwidth}
        \setlength{\tabcolsep}{5mm}{
            \begin{tabular}{@{}lc@{}}
            \toprule
            \multicolumn{2}{l}{\textbf{Counterfactual Preference Optimzation}} \\ \midrule
            Training Steps       & 7,750        \\
            Warmup Steps         & 0          \\
            Optimizer            & AdamW        \\
            Learning Rate        & 2e-5         \\
            Learning Rate Decay  & Cosine       \\
            Adam $\beta$         & (0.9, 0.98)  \\
            Weight Decay         & 0.05         \\
            Batch Size           & 4            \\ \bottomrule
            \end{tabular}
            }        
    \end{minipage}
\end{table}

While in the counterfactual preference optimization (CPO), the initial learning rate is reduced to $2\times 10^{-5}$ without the warmup. The visual encoder and text decoder are frozen out of the training. Thus, the batch size can be decreased to 4. The reinforced custom-tuning consists of 7,750 steps, executed on $2\times 2$ NVIDIA A100 GPUs. Other training parameters are the same as the fine-tuning.

\end{document}